\documentclass[letterpaper, 10 pt, journal, twoside]{IEEEtran}

\IEEEoverridecommandlockouts                              

\usepackage{amsmath,amsfonts}
\usepackage{algorithmic}
\usepackage{algorithm}
\usepackage{array}
\usepackage[caption=false,font=normalsize,labelfont=sf,textfont=sf]{subfig}
\usepackage{textcomp}
\usepackage{stfloats}
\usepackage{url}
\usepackage{verbatim}
\usepackage{graphicx}
\usepackage{amssymb}
\usepackage{mathtools}
\usepackage{leftindex}
\usepackage{tabularray}
\usepackage{xcolor}
\usepackage{gensymb}
\usepackage{indentfirst}
\usepackage{colortbl}
\usepackage{caption}
\usepackage{fancyhdr}

\usepackage{adjustbox}

\let\labelindent\relax
\usepackage{enumitem}

\definecolor{darkgray}{rgb}{0.3,0.3,0.3}
\newcommand\textbfgray[1]{\textcolor{darkgray}{\textbf{#1}}}

\UseTblrLibrary{booktabs}

\DeclarePairedDelimiter{\abs}{\lvert}{\rvert}

\makeatletter
\let\NAT@parse\undefined
\makeatother
\usepackage{hyperref}
\usepackage{cite}
\usepackage[capitalise]{cleveref}
\Crefname{equation}{Eq.}{Eqs.}
\Crefname{figure}{Fig.}{Figs.}
\Crefname{table}{Tab.}{Tabs.}
\Crefname{section}{Sec.}{Secs.}

\makeatletter
\def\ps@IEEEtitlepagestyle{%
  \def\@oddfoot{\mycopyrightnotice}%
  \def\@oddhead{\hbox{}\@IEEEheaderstyle\leftmark\hfil\thepage}\relax
  \def\@evenhead{\@IEEEheaderstyle\thepage\hfil\leftmark\hbox{}}\relax
  \def\@evenfoot{}%
}
\def\mycopyrightnotice{%
  \begin{minipage}{\textwidth}
  \centering \scriptsize
  \copyright 2024 IEEE.  Personal use of this material is permitted.  Permission from IEEE must be obtained for all other uses, in any current or future media, including reprinting/republishing this material for advertising or promotional purposes, creating new collective works, for resale or redistribution to servers or lists, or reuse of any copyrighted component of this work in other works.
  \end{minipage}
}
\makeatother

\newcommand{\insertfig}{\vspace{2pt}\includegraphics[width=0.85\linewidth]{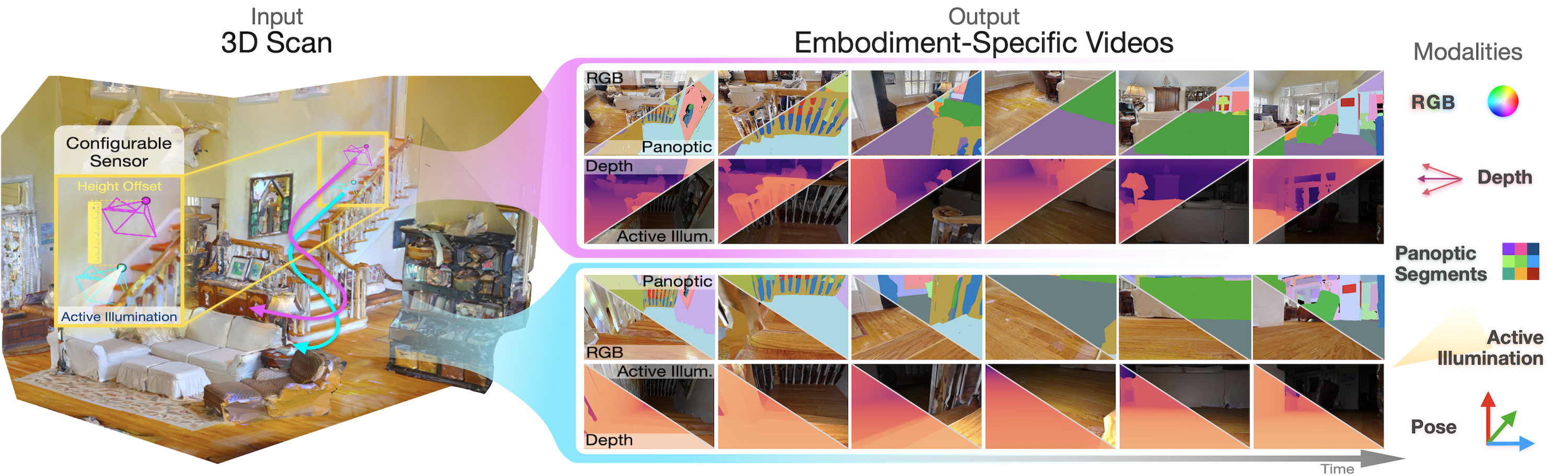}
\vspace{-4pt}
\captionof{figure}{Illustration of the data generation process used to create RGB-D videos and associated segmentation labels in the MVPd dataset. Left: Videos are collected from 3D reconstructions~\cite{yadav2023habitat} of real-world building-scale domestic environments containing clutter, heavy occlusion, and diverse objects. Right: The videos are configured to represent specific robot embodiments based on sensor placement (e.g. `tall' or `short' form factors as shown by pink and blue trajectories), sensor type, and illumination source (e.g. active or ambient).}\label{fig:teaser}\vspace{-12pt}}

\makeatletter
\apptocmd{\@maketitle}{\setcounter{figure}{0}\centering\insertfig}{}{}
\makeatother

\title{
Configurable Embodied Data Generation for\\Class-Agnostic RGB-D Video Segmentation
}

\author{Anthony Opipari$^{1}$, Aravindhan K Krishnan$^{2}$, Shreekant Gayaka$^{2}$, Min Sun$^{2}$\\Cheng-Hao Kuo$^{2}$, Arnie Sen$^{2}$, Odest Chadwicke Jenkins$^{1}$ 
\thanks{Manuscript received: June, 17, 2024; Revised September, 12, 2024; Accepted October, 7, 2024.}
\thanks{This paper was recommended for publication by Editor Markus Vincze upon evaluation of the Associate Editor and Reviewers’ comments.} 
\thanks{$^{1}$University of Michigan, {\tt\small \{topipari, ocj\}@umich.edu}}%
\thanks{$^{2}$Amazon Inc., {\tt\small \{krsar, sgayaka, minnsun, chkuo, senarnie\}@amazon.com}}%
\thanks{Digital Object Identifier (DOI): see top of this page.}
}

\begin{document}

\maketitle

\begin{abstract}

This paper presents a method for generating large-scale datasets to improve class-agnostic video segmentation across robots with different form factors.
Specifically, we consider the question of whether video segmentation models trained on generic segmentation data could be more effective for particular robot platforms {\em if} robot embodiment is factored into the data generation process.
To answer this question, a pipeline is formulated for using 3D reconstructions (e.g. from HM3DSem~\cite{yadav2023habitat}) to generate segmented videos that are configurable based on a robot's embodiment (e.g. sensor type, sensor placement, and illumination source).
A resulting massive RGB-D video panoptic segmentation dataset (MVPd) is introduced for extensive benchmarking with foundation and video segmentation models, as well as to support embodiment-focused research in video segmentation.
Our experimental findings demonstrate that using MVPd for finetuning can lead to performance improvements when transferring foundation models to certain robot embodiments, such as specific camera placements.
These experiments also show that using 3D modalities (depth images and camera pose) can lead to improvements in video segmentation accuracy and consistency.\\
Project Page: \href{https://topipari.com/projects/MVPd/}{https://topipari.com/projects/MVPd}

\end{abstract}    


\vspace{-4pt}
\section{Introduction}
\label{sec:intro}
\vspace{-2pt}

\IEEEPARstart{S}{emantic} mapping remains a critical but daunting challenge for developing robots that can understand and function autonomously in open-world human environments.
A part of this challenge stems from object-level mapping and specifically the need for robots to perceive the form and function of a vast distribution of objects that appear in domestic settings~\cite{maggio2024clio}.
Further confounding the problem are environmental factors such as occlusions, obscure lighting, affordances, and co-occurrences that increase the set of circumstances robots encounter.
Such environmental diversity is a staggering obstacle to the utility of applying data-driven object segmentation methods to mapping. This has led to \emph{increasingly large datasets} and foundation models tailored to image segmentation~\cite{kirillov2023segany,zhao2023fast}.
On the other hand, robots and their environments are both dynamic—their perspective and the position of objects in view frequently change, which motivates use of \emph{video segmentation} in-place of static image segmentation methods. 
Finally, the space of \emph{robot embodiments} is itself another factor limiting scene-understanding since training methods with data from one robot could suffer from decreased performance when transferred to a different robot with a distinct embodiment.
These observations lead to the central question addressed in this paper: \textbf{`How can roboticists scalably collect video segmentation datasets that are configurable based on their robot's specific embodiment without expensive field collection?'}

Longstanding image segmentation tasks from computer and robot vision~\cite{hariharan2014simultaneous,hafiz2020survey} have been extended to the video domain in which the goal is to accurately predict the set of pixels belonging to certain objects within each frame of a video~\cite{ren2007tracking,yang2019vis,kim2020vps}. 
In part, this is driven by the need for \emph{temporally consistent} as well as accurate segment predictions in many applications including those in robotics~\cite{zhou2022vidseg}.
This paper focuses on \emph{class-agnostic} video instance segmentation, in which algorithms are expected to predict segments for \textit{every instance of every object} and maintain temporal consistency of the predictions throughout a video.
Class-agnostic video segmentation in the open-world is especially critical for domestic robots that operate in cluttered home environments containing a wide distribution of object categories, not all of which can be included in finite training datasets.
A natural consequence of moving to the video domain is an associated increase in the expense of collecting videos with densely annotated segmentation masks~\cite{yang2019vis,miao2022vipseg}.
Thus, in contrast to the image domain, for which many large-scale datasets have been published~\cite{gupta2019lvis,OpenImagesSegmentation,kirillov2023segany}, there are relatively few comparably sized video segmentation datasets.

This work sets out to provide a scalable solution for creating video segmentation datasets that are customised to specific robot embodiments.
A key insight inspring the work is the increasing availability of 3D reconstructions~\cite{xiao2013sund3d,Matterport3D,yadav2023habitat} of real-world, cluttered, domestic environments and \emph{the potential for these representations to be used to control for robot embodiment while generating large-scale and densely annotated segmentation videos}.
With this insight, the present paper makes the following contributions:
\begin{enumerate}
    \item The {\bf Massive Video Panoptic dataset (MVPd)} and data generation pipeline for configurable robot video segmentation data. In total, MVPd is \textit{more than 45$\times$ larger} than existing video segmentation benchmarks with \textbf{18K annotated RGB-D videos, $>$6M images, and 162M masks}. The data generation pipeline can create large-scale simulated datasets controlling for embodiment such as sensor placement and scene illumination.
    \item Benchmarking experiments are conducted to evaluate state-of-the-art algorithm performance using MVPd on the class-agnostic video instance segmentation task. Results are included to compare both foundation models and specially designed video segmentation models.
    \item Extensive ablation experiments are carried out to establish the impact of sensor placement on segmentation quality as well as the potential for 3D data, in the form of depth images and camera pose, to improve segmentation consistency. Presented findings show that {\bf configuring for embodiment during the training data generation process can improve segmentation quality and consistency across distinct robot embodiments}.
\end{enumerate}

\vspace{-4pt}
\section{Related Work}
\label{sec:relatedwork}

\textbf{Video segmentation} can be summarized into two tracks: a semantic track and a class-agnostic track.
Within the semantic track, video semantic segmentation expects pixel-level segments for each object in each video frame along with a label for their semantic categories while video instance segmentation distinguishes between multiple objects of the same category~\cite{yang2019vis}. 
Video panoptic segmentation considers cases in which certain categories are nebulous in shape such as the sky, floor, or `stuff'~\cite{kim2020vps}.
Most methods proposed for the semantic track specialize to specific semantic categories~\cite{vertens2017smsnet,li2022videoknet,miao2022vipseg,yang2022tevit,xu2023viposeg,li2023tube}.
For instance, Video K-Net~\cite{li2022videoknet} learns a kernel for each class and differentiates among the kernels of distinct instances.
Similarly, transformer-based architectures such as PAOT~\cite{xu2023viposeg} and Tube-Link~\cite{li2023tube} have been proposed to associate features across images based on which category each feature belongs to.
When algorithms are expected to segment objects \textit{regardless of their semantic class}, \textit{class-agnostic} video segmentation is considered~\cite{siamv2021cas}.
A special case is video object segmentation~\cite{ren2007tracking}, in which only one or a few specific objects of interest, chosen by their foreground position~\cite{tokmakov2017vos} or manual masking~\cite{perazzi2017vos,caelles2017vos}, are to be segmented.
Within class-agnostic video segmentation, much of the work uses motion cues like optical flow to separate each object~\cite{cheng2017segflow,siamv2021cas}.
Many approaches consider only video object segmentation, where foreground motion is feature-rich, and hence do not segment \textit{every} object in the scene. 
For example, SegGPT~\cite{wang2023seggpt} segments a specific object of interest using an image-level foundation model and in-context coloring.
For segmenting all objects in the autonomous vehicle setting, Siam et al.~\cite{siamv2021cas} propose using object motion estimation.
Inspired by the work to use optical flow as a feature, the present paper considers whether a robot's egocentric motion (i.e. pose estimated by odometry), can be used as a feature within class-agnostic segmentation.

\textbf{Open-world segmentation} has been a growing topic of interest for image-level segmentation~\cite{bucher2019zssem,danielczuk2019segmenting,zheng2021zsis,xiang2021learning,ornek2023super}.
A few recent works have carried the topic into video segmentation~\cite{cheng2017segflow,du2021unseenvidseg,siamv2021cas,xu2023viposeg,nunes2022opensegreview,cheng2023tracking}.
Again, optical flow and object motion estimation have been used as features to distinguish foreground and background objects in zero-shot applications~\cite{cheng2017segflow,du2021unseenvidseg,siamv2021cas}.
More recently, Xu et al.~\cite{xu2023viposeg} demonstrate the challenge for generalizing video segmentation to unseen categories in videos collected from YouTube.
Wang et al.~\cite{wang2023towards,wang2024ov} and Li et al.~\cite{li2024omg} propose using language features to extend video segmentation models to the open-world.
In contrast, we set out to explore the potential for open-world video segmentation in domestic environments by embodied robots.
We are inspired by recent foundation models tailored to zero-shot transfer~\cite{caron2021dino,kirillov2023segany}.
In particular, promptable models such as Segment Anything Model (SAM)~\cite{kirillov2023segany} and FastSAM~\cite{zhao2023fast} have led to a number of follow-on works demonstrating their potential as transferrable models.
Cen et al.~\cite{cen2023segment} showed how recursive prompting to SAM with neural radiance fields can be used to extract volumetric segments from static scenes.
We take inspiration from this work and consider how promptable segmentation can be used recursively \textit{over time} in class-agnostic video segmentation.

\begin{table*}[t]
\centering  
\resizebox{0.9\textwidth}{!}{
\begin{tblr}{
        colspec={Q[l]Q[l,m]lQ[c,m,10mm]rrQ[r,m,10mm]rQ[c,m,12mm]Q[c,m,10mm]Q[c,m,12mm]},
    }
    \toprule
    Dataset & Setting & Modality & Camera Pose & Videos & Images & Obj. per Video & Classes & Stuff Categories & Unseen Split & Sensor Placement Control\\
    \midrule
    Cityscapes VPS\cite{kim2020vps} & Roads & RGB-D\textsuperscript{\textdagger} & $\times$ & 500 & 3,000 & 28.79\textsuperscript{*} & 19 & \checkmark & $\times$ & $\times$\\
    KITTI-STEP\cite{weber2021step} & Roads & RGB & $\times$ & 50 & 19,103 & 53.76\textsuperscript{*} & 19 & \checkmark & $\times$ & $\times$\\
    MOTC-STEP\cite{weber2021step} & Pedways & RGB & $\times$ & 4 & 2,075 & 38.00\textsuperscript{*} & 7 & \checkmark & $\times$ & $\times$\\
    YTVIS\cite{yang2022vis} & Wild & RGB & $\times$ & 4,046 & 128,930 & 2.10\textsuperscript{*} & 40 & $\times$ & $\times$ & $\times$\\
    OVIS\cite{qi2022ovis} & Wild & RGB & $\times$ & 901 & 62,641 & 5.90\textsuperscript{*} & 25 & $\times$ & $\times$ & $\times$\\
    VIPSeg\cite{miao2022vipseg} & Wild & RGB & $\times$ & 3,536 & 84,750 & 13.65\textsuperscript{*} & 124 & $\checkmark$ & $\times$ & $\times$\\
    VIPOSeg\cite{xu2023viposeg} & Wild & RGB & $\times$ & 3,149 & 75,022 & 13.65\textsuperscript{*} & 125 & \checkmark & \checkmark & $\times$\\
    LV-VIS\cite{wang2023towards} & Wild & RGB & $\times$ & 4,828 & 111,298 & 5.3\phantom{\textsuperscript{*}} & \textbf{1196} & \checkmark & \checkmark & $\times$ \\
    $\text{UVO}_{\mathcal{D}}$\cite{Wang_2021_ICCV} & Kinetics & RGB & $\times$ & 1,017 & 91,530 & 13.52\phantom{\textsuperscript{*}} & 1 & $\times$ & \checkmark & $\times$ \\
    MVPd & Domestic & RGB-D & \checkmark & \textbf{18,000} & \textbf{6,055,628} & \textbf{94.39}\phantom{\textsuperscript{*}} & 40 & \checkmark & \checkmark & \checkmark\\
    \bottomrule
\end{tblr}
}
\vspace{-4pt}
\caption{Comparison of related video instance and panoptic segmentation benchmarks with MVPd. {}\textsuperscript{\textdagger}Noisy depth can be computed from stereo images in Cityscapes dataset~\cite{Cordts2016Cityscapes}. MVPd provides dense depth as measured by a Matterport scanner. {}\textsuperscript{*}Denotes calculation based only on publicly available data (i.e. not including private evaluation data).}
\label{tab:mvpd}
\vspace{-16pt}
\end{table*}

\textbf{Large-scale datasets} for video segmentation are expensive to collect and annotate, which has been identified as a limiting factor for learning-based solutions~\cite{yang2019vis,miao2022vipseg}.
Despite these challenges, a number of benchmark datasets have been introduced~\cite{kim2020vps,weber2021step,yang2022vis,qi2022ovis,miao2022vipseg,xu2023viposeg,Wang_2021_ICCV} and are summarized in ~\cref{tab:mvpd}.
For exterior road settings, Weber et al.~\cite{weber2021step} introduced KITTI-STEP and MOTChallenge-STEP, totaling $>$21K annotated images.
Beyond autonomous driving, datasets collected `in the wild' have led to even larger video segmentation datasets.
Yang et al.~\cite{yang2019vis} used videos from YouTube for video instance and object segmentation tasks.
More recently, Miao et al.~\cite{miao2021vspw} introduced the VSPW dataset for video semantic segmentation before adding additional labels to support video panoptic segmentation in VIPSeg~\cite{miao2022vipseg} and zero-shot objects by Xu et al.~\cite{xu2023viposeg}.
For class-agnostic video segmentation, Wang et al.~\cite{Wang_2021_ICCV} introduced the $\text{UVO}_{\mathcal{D}}$ dataset which focuses its annotations on kinetic human-object interactions.
In contrast to the existing video segmentation benchmarks, MVPd provides 3D input modalities in the form of ground truth depth images and camera pose for every RGB image.
In addition, MVPd focuses on indoor domestic environments as opposed to exterior settings~\cite{kim2020vps,weber2021step}, or videos from the wild~\cite{yang2022vis,qi2022ovis,miao2022vipseg,xu2023viposeg,Wang_2021_ICCV}.
The substantial object clutter observed in MVPd (94.39 objects per video) suggests it better captures unstructured domestic environments than existing video segmentation benchmarks.
A separate line of work has used increasingly available 3D scanners to collect and annotate 3D point and mesh-based datasets from real-world spaces~\cite{xiao2013sund3d,Matterport3D,yadav2023habitat,dai2017scannet}, for use in 3D segmentation benchmarks.
Using 3D mesh datasets, Eftekhar et al.~\cite{eftekhar2021omnidata} introduced a pipeline to create `steerable' datasets for specific computer vision tasks and demonstrated their benefit on non-embodied image-based tasks.
Building upon these ideas, the MVPd data generation pipeline focuses on video segmentation specifically and controlling for the features of embodiment (sensor type, sensor placement and illumination sources) that can impact video segmentation consistency and accuracy for downstream domestic robots.

\begin{figure*}[b]
\vspace{-8pt}
\centering    \includegraphics[width=0.85\textwidth]{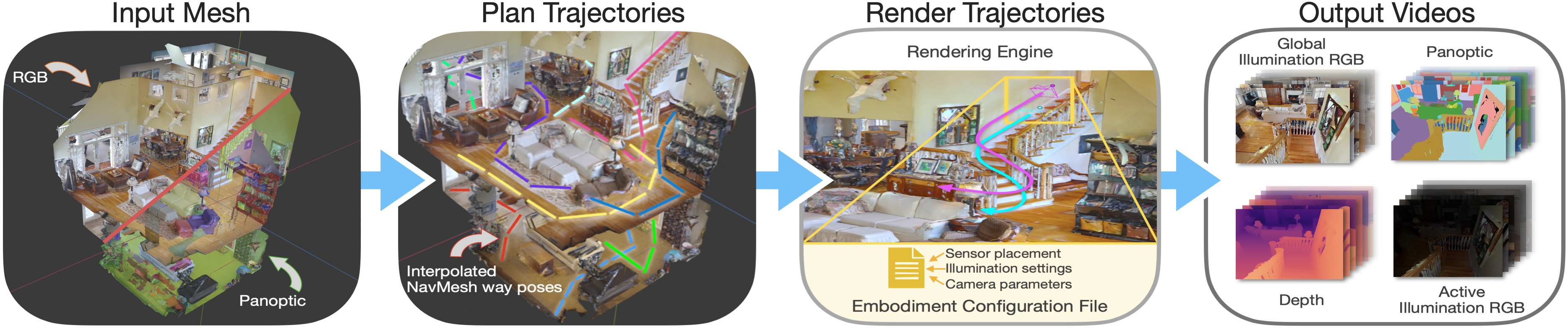}
\vspace{-4pt}
\captionof{figure}{Illustration of the data generation pipeline used to create MVPd. Left: Using input meshes that contain RGB and segmentation textures, the pipeline generates sparse random paths with a collision-free NavMesh planner~\cite{habitat19iccv} and interpolates them into dense trajectories of way poses. Right: The motion trajectories are refined according to an embodiment configuration file and rendered to output videos.}
    \label{fig:data_pipeline}
    \vspace{-16pt}
\end{figure*}

\section{MVPd: Massive Video Panoptic Dataset}

\vspace{-4pt}

MVPd is introduced to support research on embodied class-agnostic video instance segmentation and the potential for 3D modalities to benefit video segmentation algorithms.
In total, MVPd contains 18,000 densely annotated RGB-D videos, 6,055,628 individual image frames with ground truth 6DoF pose, and 162,115,039 masks.
Each video in MVPd contains between 100 and 600 image frames rendered at 640x480 resolution.
Videos are captured from scenes of HM3DSem~\cite{yadav2023habitat}, containing real-world building-scale domestic environments such as homes, offices, and retail spaces.
Each scene is on the scale of a multi-floor building with on average \textgreater14 rooms per scene, and ~60 objects per room~\cite{yadav2023habitat}. This is the largest semantically annotated indoor scene dataset we are aware of, covering \textgreater20,000m$^2$ of navigable area and \textgreater2x the number of unique object instances as are available in comparable scene datasets.
Annotated segments are assigned to one of 40 Matterport categories~\cite{Matterport3D}.

\vspace{-4pt}
\subsection{Data Generation Pipeline}
\vspace{-2pt}

The key insights inspiring MVPd's data generation pipeline are that (1) \textit{instance annotations at the mesh-level enable inexpensive segment annotations at the video-level} and (2) \textit{the mesh representation for scenes enables embodiment-specific configuration for each video at scale}.
As a source of real-world scenes, the HM3DSem dataset~\cite{yadav2023habitat} provides mesh representations of $216$ real-world building-scale environments created with a Matterport scanner.
Alternative scene datasets, such as Matterport3D~\cite{Matterport3D}, are compatible with our data generation pipeline and offer an added source of annotated scenes that could be rendered as videos alongside MVPd in the future.

\begin{figure}[t]
\vspace{-10pt}
\centering    
\includegraphics[width=0.9\columnwidth,]{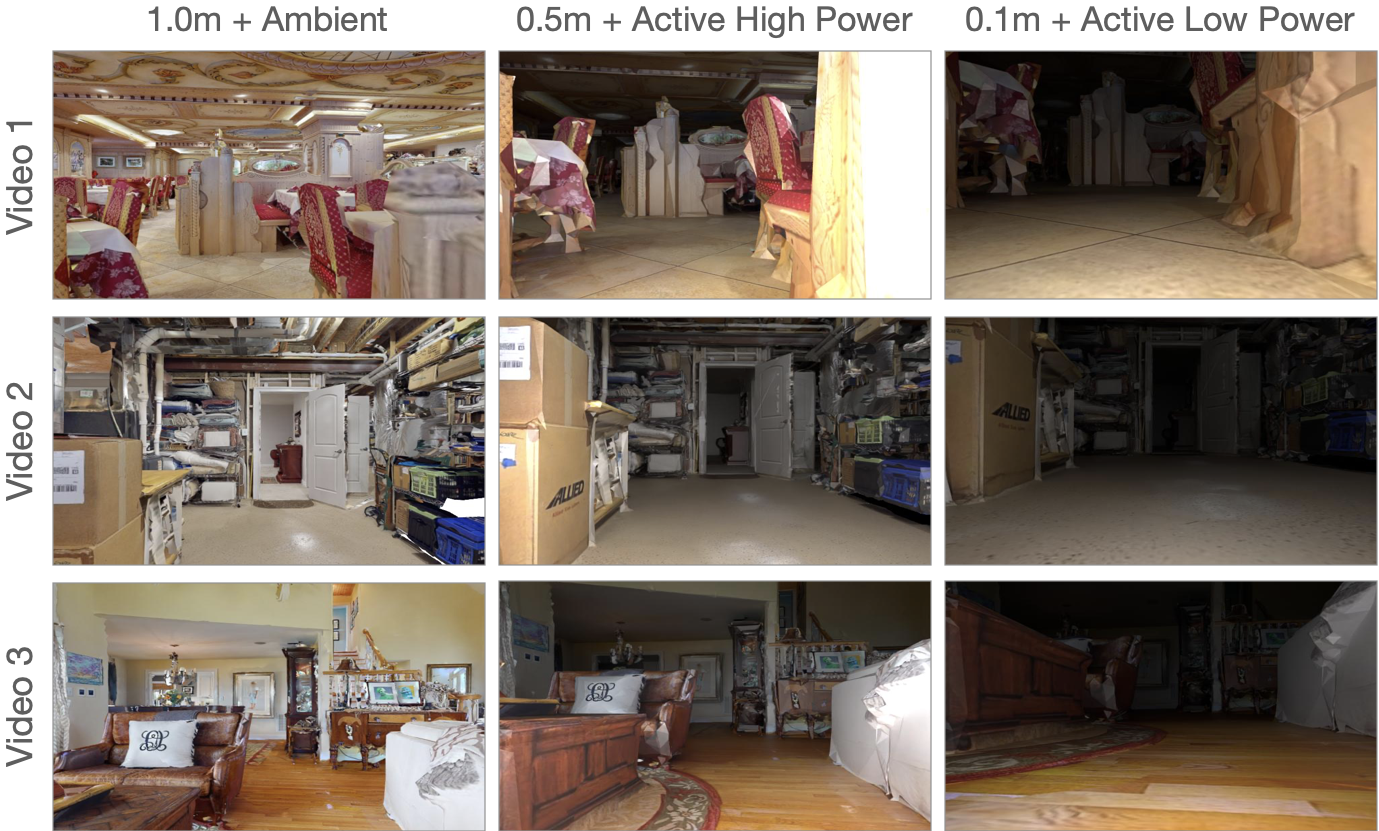}
\vspace{-4pt}
\captionof{figure}{Impact of specific embodiments (sensor placement \& active illumination) on visual features rendered by the MVPd}
    \label{fig:pipeline_examples}
    \vspace{-12pt}
\end{figure}

As illustrated in~\cref{fig:data_pipeline}, the data generation pipeline takes input meshes that contain RGB color and segmentation textures. 
Using the mesh, random paths of sparse waypoints are planned using a collision-free NavMesh planner from Habitat-Sim~\cite{habitat19iccv} which are then interpolated by the pipeline to form trajectory plans of smooth and dense wayposes.
Specifically, the pipeline uses linear interpolation to ensure the trajectories are smooth to within 5cm of linear displacement and 0.5\degree of axis-angle rotation.
Next, an embodiment configuration file is used to refine the trajectory before rendering.
According to the embodiment specification, an RGB-D camera is placed at each waypoint pose to render a corresponding video frame that may include RGB, depth, and panoptic labels.
The embodiment specification may control for sensor placement and illumination source (i.e. ambient or active) and power (Watts) as shown in~\cref{fig:pipeline_examples}.

MVPd videos are rendered as follows:
180 scenes from HM3DSem are chosen for inclusion based on public availability.
For each scene, 50 random start and end waypoint pairs are chosen without replacement for path planning, resulting in 50 trajectories per scene.
For each trajectory, embodiment specification is set at both 1m and 0.1m above the floor to emulate challenging perspectives taken by home robots. 
Thus, 100 videos are rendered for each of the 180 scenes with an average trajectory distance of 7.48m.

\subsection{Zero-Shot Subset: Seen and Unseen Categories}
\label{sec:zsmvpd}
\vspace{-2pt}

Zero-shot learning refers to the application of machine learning models on data categories not seen during training (the `open-world')~\cite{bucher2019zssem}.
It is especially relevant for robots deployed in the real-world and it encompasses any learning-based task including detection, segmentation, \textit{and} video segmentation.
A zero-shot subset of MVPd is included to support research on generalizable video segmentation.
Evaluations on the zero-shot subset are intended to reflect an expectation of model performance in the open-world.

The zero-shot subset of MVPd is defined by a set of `\textit{seen}' and `\textit{unseen}' object classes.
Only the seen class are available to models during training, while both the seen and unseen classes are used for testing.
To select these classes, we follow a process similar to~\cite{bansal2018zero,ornek2023super}.
All object instances within MVPd belonging to the `Misc' and `Objects' Matterport categories are considered for the unseen class since these encompass a broad collection of instance morphologies unlike the remaining 38 Matterport super-categories.
Next, for each object instance a CLIP-embedding (ViT-L/14)~\cite{radford2021clip} is generated using the instance's corresponding human-annotated text description.
The embeddings are then clustered by the $k$-means algorithm with 20 clusters.
20\% of the clusters are assigned to the unseen set and the remaining clusters to the seen set. The split of clusters is based on the number of training videos that observe objects from the various cluster, with the least frequently observed clusters chosen for the unseen class.
As illustrated in ~\cref{fig:mvpd_zeroshothist}, the unseen class include objects relating to clothes storage (e.g. closet, shelf, cubby), hobby items (e.g. pianos, aprons, aquariums), stands (e.g. book rack, computer tower, foot stand), and soap (e.g. detergent, washing powder).
Videos containing any object from the zero-shot unseen class are excluded from the training set to create a zero-shot subset.

\section{Class-Agnostic Video Instance Segmentation}
\label{sec:caVIS}

\vspace{-2pt}
\noindent\textbf{\underline{Problem Definition:}}
Given a video sequence consisting of $T$ frames, consider a temporal window of $k\leq T$ consecutive frames denoted by $I^{t:t+k} = \{I^t, I^{t+1}, \ldots, I^{t+k}\}$.
Following the definition of VPS~\cite{kim2020vps}, a tube prediction corresponding to the $k$-span window is defined as a track of frame-level segments, $\hat{u}_{z_i}=\{\hat{s}^t,\ldots,\hat{s}^{t+k}\}_{z_i}$ where $z_i$ represents a unique instance identifier.
Ground truth segment tubes are defined analogously using annotated segments at each frame in the window.
The goal of class-agnostic video instance segmentation is to accurately segment every instance of every object within a video, regardless of the objects' semantic categories.
Unlike VPS~\cite{kim2020vps} and VIS~\cite{yang2019vis}, class-agnostic video instance segmentation does not require predicted tubes be assigned to one of a predefined set of semantic classes.

\noindent\textbf{\underline{Evaluation Metric:}}
\label{sec:metric}
A growing set of foundation segmentation models including SAM~\cite{kirillov2023segany} and FastSAM~\cite{zhao2023fast} have been proposed for broad applicability and this paper set out to include them in its evaluations.
However, because these models produce overlapping predictions they cannot be directly evaluated by the video panoptic quality metric introduced by Kim et al.~\cite{kim2020vps} as it requires no-overlap between predicted segments~\cite{kirillov2019panoptic}.
To build upon the video panoptic quality metric, we propose a slight modification to enable its evaluation of models regardless of their predictions' overlap. 
For this modification we borrow inspiration from class-agnostic instance segmentation metrics used for images~\cite{davesam2019,ornek2023super} to define class-agnostic Video Instance Segmentation Quality (VSQ) as follows:

\begin{figure}[t!]
  \centering
    \includegraphics[width=0.9\linewidth]{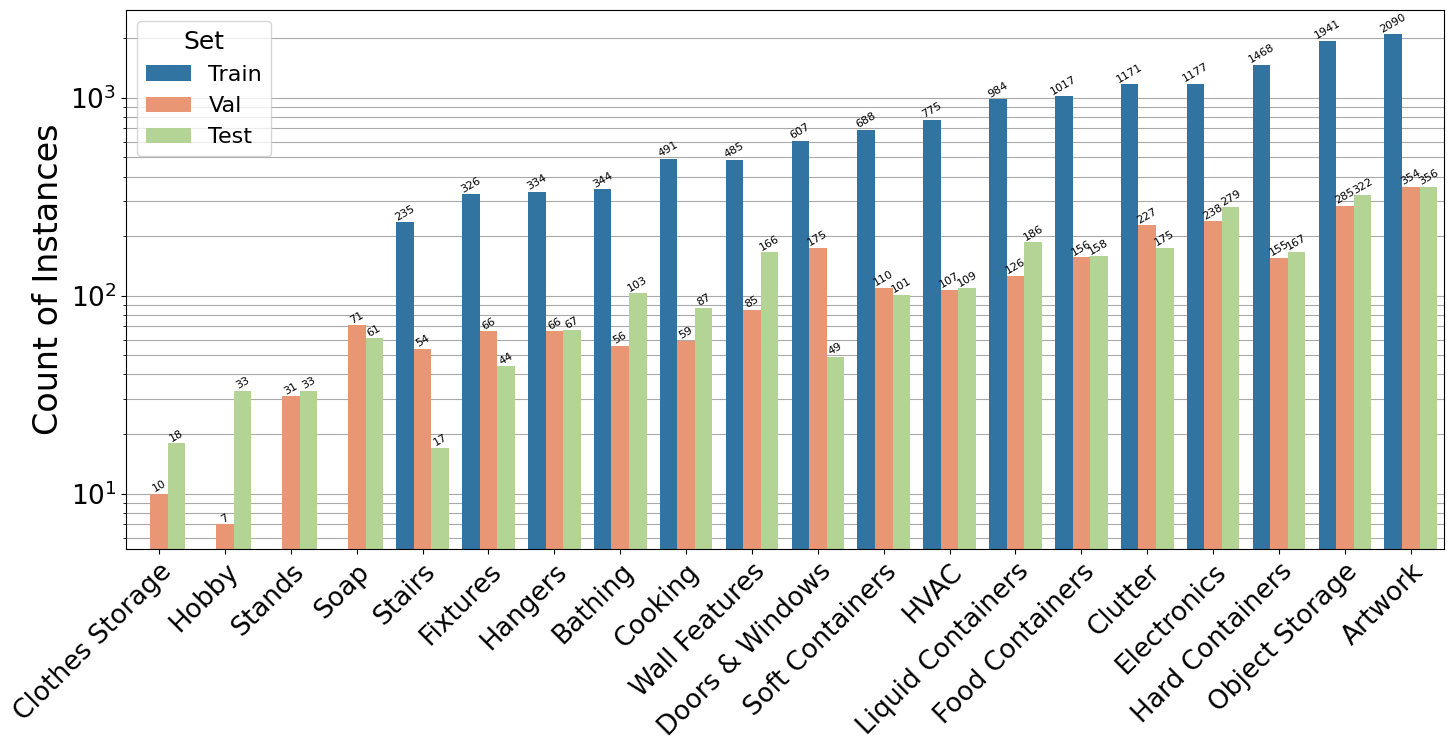}
    \vspace{-6pt}
    \caption{Distribution of the 20 object clusters used for creating the zero-shot subset of MVPd. Object instances from the 'Objects' and 'Misc' super-categories in MVPd, as defined in~\cite{Matterport3D}, are grouped into 20 clusters using $k$-means with the CLIP-embedding~\cite{radford2021clip} of each instance's human-annotated text description. Clusters are then ranked by the number of associated training videos, and the smallest 20\% (clothes storage, hobby items, stands, and soap) are defined as the zero-shot classes. The text summary of each cluster (e.g. Clothes Storage, Hobby, etc.) are defined based on manual inspection.}
    \label{fig:mvpd_zeroshothist}
    \vspace{-12pt}
\end{figure}

\begin{figure*}[t!]
  \centering
   \includegraphics[width=0.85\textwidth]{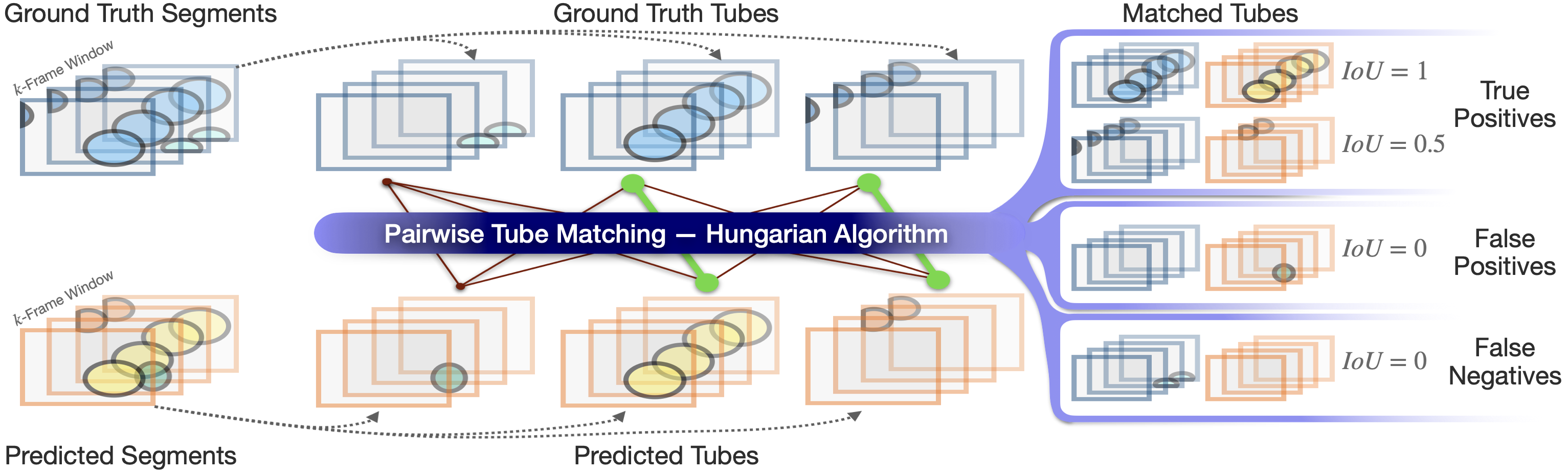}
    \vspace{-4pt}
   \caption{Illustration of the the class-agnostic video segmentation quality (VSQ) metric. Given a fixed $k$-frame window size, ground truth and predicted segments are isolated into segment tubes and then matched using an optimal assignment algorithm according to pairwise F-measure. Once matched, the set of true positive, false positive and false negative tubes are counted and a per-match $IoU$ is computed.}
   \vspace{-16pt}
   \label{fig:VSQ}
\end{figure*}

For a fixed $k$-frame window size, VSQ$^k$ is computed by measuring the overlap between each temporally aligned ground truth and predicted segment tube of length $k$.
As illustrated in ~\cref{fig:VSQ}, VSQ$^k$ is computed by first matching each ground truth tube to each predicted tube using the Hungarian algorithm.
The resulting assignment maximizes the sum total F-measure over each tube match.
Each matched tube is considered a true positive (TP).
Any predicted tube that is left unmatched is considered a false positive (FP) while any unmatched ground truth tube is considered a false negative (FN).
Using the pairwise intersection over union (IoU) between each TP, VSQ$^k$ is calculated as follows:

\vspace{-6pt}
\begin{equation}
  \text{VSQ}^k = \frac{\sum_{(u,\hat{u})\in TP} IoU(u,\hat{u})}{\abs{TP}+\frac{1}{2}\abs{FP}+\frac{1}{2}\abs{FN}}
  \label{eq:VSQ^k}
\end{equation}
\vspace{-8pt}

As for video panoptic quality~\cite{kim2020vps}, a final VSQ score is computed by averaging VSQ$^k$ over a set of windows, $K$:

\vspace{-6pt}
\begin{equation}
  \text{VSQ} = \frac{1}{K}\sum_k \text{VSQ}^k
  \label{eq:caVSQ}
\end{equation}
\vspace{-8pt}

For all experiments in this paper, $K=\{1,5,10,15\}$ and VSQ$^k$ is calculated using a window stride of $15$ frames. 
To enable evaluation of frame-level models by VSQ, models may either directly output tubes or have output tubes formatted from frame-level predictions before evaluation. To format tubes from frame-level predictions, the Hungarian algorithm is applied in a pairwise fashion over their frame-level output segments. For models in this paper's experiments, Video K-Net and Tube-Link output tubes directly while SAM, FastSAM and FastSPAM have tubes formatted as described.

\section{FastSPAM: Self-Prompting Anything Model}
\label{sec:fastspam}

Building on recent foundation models for promptable segmentation~\cite{kirillov2023segany,zhao2023fast}, we develop a self-prompting mechanism to enhance FastSAM~\cite{zhao2023fast} for greater accuracy and consistency in class-agnostic video instance segmentation.
The key insight for using self-prompting lies in the hypothesis that promptable segmentation models like FastSAM can be made to output segments with reduced flickering if the prompts fed as input are informed by 3D spatial cues.
To investigate this hypothesis, spatio-temporal self-prompting was developed to ensure the set of input prompts remain grounded in 3D space regardless of changes in camera viewpoint, thereby aiming to reduce the amount of flickering in the model's output over sequential frames.
The resulting Fast Self-Prompting Anything model is referred to as FastSPAM. 
For a visual illustration, readers are referred to the supplementary video.

Fast\textit{SPAM} performs segmentation in two stages: (1) an all-instance stage, which detects and predicts segments given an image as input. (2) A prompt-guided selection stage, which uses the input image together with a prompt (i.e. a point coordinate) to refine the detected segments.
FastSPAM uses a sequence of RGB-D images, $(I^{0:T},D^{0:T})$, as well as the corresponding camera projection matrices, $C^{0:T}$, as input.
The camera projection matrix is defined to include both intrinsic and extrinsic parameters, $C=K\begin{bmatrix}R\vert T\end{bmatrix}$, where $K$ is the camera calibration matrix and $\begin{bmatrix}R\vert T\end{bmatrix}$ is a homogeneous matrix defining the camera pose in world coordinate frame.

At each point in time FastSPAM maintains a set of self-prompts $P^t=\{p_0,\ldots,p_N\}^t$ where each self-prompt $p^t_i\in\mathbb{R}^3$ is a 3D point in the world coordinate frame describing the estimated centroid of each object.
Given these pieces of information at time, $t-1$, FastSPAM first applies the YOLOv8-seg~\cite{jocher2023YOLO} method on the image $I^t$ to perform all-instance segmentation.
Next, the self-prompts are projected into the current image frame: $P'=\{C^{t}p^{t-1}_0,\ldots,C^{t}p^{t-1}_N\}$.
Predicted segments from the all-instance stage are merged by union according to the self-prompts to ensure a single mask is predicted for each self-prompt.
Finally for future predictions, an updated set of self-prompts, $P^{t}$, is calculated by converting (or `unprojecting') the pixel-coordinate of each predicted segment's centroid into a 3D point in the world coordinate frame using the depth and camera matrix.

\section{Experiments}
\label{sec:exp}

\label{sec:exp_setup}
\vspace{-4pt}

\noindent\textbf{\underline{Dataset:}}
MVPd is split into training, validation, and test subsets randomly at the scene-level (i.e. scenes in the training set have no videos represented in the test set).
The held-out test set is used for all evaluations, while the training and validation sets are used for model learning and tuning.

\noindent\textbf{\underline{Baseline Models:}}
Two types of baseline models are considered: foundation models for image segmentation and finetuned models for video instance segmentation.
The specific foundation models used for this paper include Segment Anything Model (SAM)~\cite{kirillov2023segany} and FastSAM~\cite{zhao2023fast}.
Two variants of SAM are used: SAM-B which uses ViT-B as its backbone and SAM-H which uses ViT-H as its backbone.
In contrast, Video K-Net~\cite{li2022videoknet}, Tube-Link~\cite{li2023tube}, and OV2Seg~\cite{wang2023towards} are used as baselines that were developed for video instance segmentation.
Two variants of each baseline are included in these experiments: one variant uses a ResNet50 backbone and another uses a Swin-base or Swin-large backbone.

\noindent\textbf{\underline{Implementation Details:}}
Pre-trained SAM models are evaluated in automatic mode using default settings (32x32 uniform grid of point prompts).
Finetuned FastSAM and FastSPAM models are trained on a single RTX A6000 GPU using a batch size of 16 images for 169K iterations (1 MVPd epoch).
All other hyperparemters of FastSAM are left unchanged.
Video K-Net and Tube-Link are trained with a mini-batch of 1 sample per GPU and a frame-range of 5 images per sample. 
OV2Seg models (ResNet50, Swin-base) are trained with a batch size of 16 and 8 images respectively.
Each baseline is trained in a distributed fashion with 8 Tesla V100-GPUs until convergence (200K iterations).
Otherwise, the baseline models are trained using the respective authors' published implementations and hyperparameter and pre-training settings; we did not tune the hyperparameters used for these models except to increase the total training iterations to ensure model convergence on MVPd (200k iterations instead of ~100k~\cite{li2022videoknet} and 6-8k~\cite{li2023tube}).
All models are trained with instance labels.

\begin{table}[b]
\vspace{-8pt}
  \centering
  \resizebox{0.9\columnwidth}{!}{
\begin{tblr}{
        colspec={Q[l]Q[c]Q[c]Q[c]Q[c]Q[c]Q[c,m]},
        cell{1}{1} = {r=2}{m},
        cell{1}{2} = {r=2}{m},
        cell{1}{7} = {r=2}{m}
    }
    \toprule
     Method & Backbone & \SetCell[c=4]{c}{{{VSQ$^k$ with $k$-Frame Window}}} & & & & VSQ\\
     \cmidrule[lr]{3-6}
    & & $k=1$ & $k=5$ & $k=10$ & $k=15$ & \\
    \midrule
    SAM & ViT-H & 29.78 & 19.33 & 13.38 & 10.27 & 18.19 \\
    SAM & ViT-B & 32.48 & 20.48 & 13.99 & 10.70 & 19.41 \\
    FastSAM & YOLOv8 & \textbf{41.02} & \textbf{31.13} & \textbf{24.19} & \textbf{20.01} & \textbf{29.09} \\
    \bottomrule
\end{tblr}
}
\vspace{-4pt}
  \caption{Evaluating pre-trained foundation models (SAM-H, SAM-B, FastSAM) on the class-agnostic video instance segmentation task. Evaluation performed on videos from the held-out test set of MVPd and the VSQ evaluation metric (\cref{sec:metric}).}
  \label{tab:exp_pretrained}
\end{table}

\vspace{-6pt}
\subsection{Pre-trained Foundation Models}
\label{sec:exp_transfer}
\vspace{-2pt}

In the first experiment, we set out to understand the effectiveness of foundation models that were pre-trained on a large-scale class-agnostic image segmentation dataset (SA-1B~\cite{kirillov2023segany}) when evaluated on MVPd and class-agnostic \textit{video} instance segmentation.
To address this question, pre-trained foundation models~\cite{kirillov2023segany,zhao2023fast} (ViT-H SAM, ViT-B SAM), and FastSAM) were applied to each image in MVPd's test set and evaluated using the VSQ metric (\cref{sec:metric}).
Video K-Net and Tube-Link are excluded from this evaluation since the pre-trained parameters publicly available are not intended for zero-shot transfer.
The quantitative results are included in ~\cref{tab:exp_pretrained}.
Comparing each model on the VSQ metric suggests an inverse relationship between model complexity and VSQ score.
Inspection suggests that SAM's VSQ score suffers due to SAM's frequent sub-part predictions, which are considered to be false-positives in this task.
\textbf{These results indicate that directly transferring image-based foundation models to video instance segmentation, despite their large-scale pre-training dataset, results in limited segmentation quality.}

\begin{figure}[t]

  \centering
        \includegraphics[width=0.9\columnwidth]{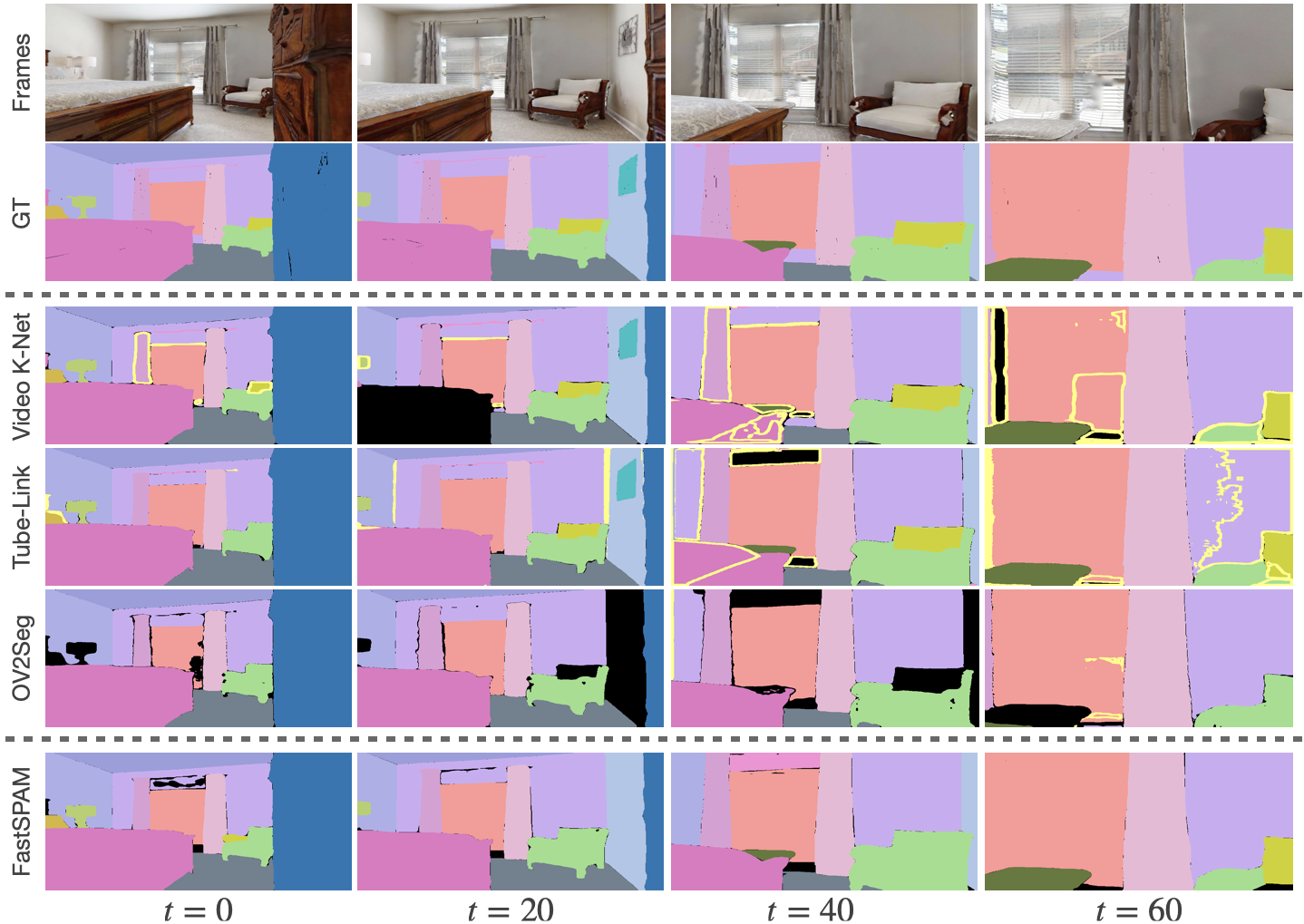}
    \vspace{-6pt}
   \caption{Qualitative results comparing the swin-variants of Video K-Net, Tube-Link, OV2Seg, and FastSPAM. FastSPAM exhibits segments that are more consistent and accurate than the baselines and with fewer false positive predictions (outlined in neon yellow).}
    \vspace{-14pt}
    \label{fig:qual_all}
\end{figure}

\vspace{-6pt}
\subsection{After Finetuning Models}
\label{sec:exp_finetuning}
\vspace{-2pt}

Next, we set out to evaluate the extent to which MVPd can support improved segmentation quality via finetuning.
FastSAM , Video K-Net, and Tube-Link were finetuned on the MVPd training set and evaluated on its test set.
The SAM models are excluded from this experiment due to resource limitations and their substantial training GPU requirements.

Quantitative results are shown in~\cref{tab:exp_finetuned} indicating that for each window setting considered, FastSAM achieved the highest VSQ$^k$ as well as the highest VSQ score of 49.17\%, which is +6.65\%, +24.05\%, and +10.98\% higher than top-performing baseline models respectively.
The relative VSQ performance difference between Video K-Net, OV2Seg, and Tube-Link may result from Tube-Link's more complicated attention-based linking.
Qualitative examples comparing each model are included in~\cref{fig:qual_all}.
\textbf{These results suggest image-based foundation models which were pre-trained on large-scale image datasets can be competitive with models designed specifically for video segmentation, if given sufficient in-domain data for finetuning.}

\begin{table}[b]
  \centering
  \resizebox{0.95\columnwidth}{!}{
\begin{tblr}{
        colspec={Q[l]Q[c]Q[c]Q[c]Q[c]Q[c]Q[c,m]},
        cell{1}{1} = {r=2}{m},
        cell{1}{2} = {r=2}{m},
        cell{1}{7} = {r=2}{m}
    }
    \toprule
     Method & Backbone &\SetCell[c=4]{c}{{{VSQ$^k$ with $k$-Frame Window}}} & & & & VSQ\\
     \cmidrule[lr]{3-6}
    & & $k=1$ & $k=5$ & $k=10$ & $k=15$ & \\
    \midrule
    Video K-Net & ResNet50 & 49.65 & 49.60 & 38.10 & 29.56 & 41.73 \\
    Video K-Net & Swin-base & 50.83 & 50.50 & 38.71 & 30.05 & 42.52 \\
    Tube-Link & ResNet50 & 45.13 & 20.50 & 16.08 & 14.05 & 23.94 \\
    Tube-Link & Swin-large & 48.17 & 21.14 & 16.60 & 14.55 & 25.12 \\
    OV2Seg & ResNet50 & 39.96 & 38.52 & 37.51 & 36.71 & 38.18 \\
    OV2Seg & Swin-base & 40.22 & 38.58 & 37.43 & 36.52 & 38.19 \\
    FastSAM & YOLOv8 & \textbf{61.18} & \textbf{52.03} & \textbf{44.38} & \textbf{39.09} & \textbf{49.17} \\
    \bottomrule
\end{tblr}
}
\vspace{-4pt}
  \caption{Evaluating the impact of finetuning on model performance under VSQ metric and the MVPd test set.}
  \label{tab:exp_finetuned}
  \vspace{-12pt}
\end{table}

\vspace{-6pt}
\subsection{Using Depth and Camera Pose Modalities}
\label{sec:exp_depth_pose}
\vspace{-2pt}
We next set out to understand whether 3D modalities can be used to improve model segmentation quality.
To answer this question, a self-prompting mechanism and FastSPAM model (\cref{sec:fastspam}) were developed that use camera pose and depth (which FastSAM does not use) to recursively generate the prompts used to create its segment predictions. 

\begin{figure}[t!]
  \centering
    \includegraphics[width=0.85\columnwidth]{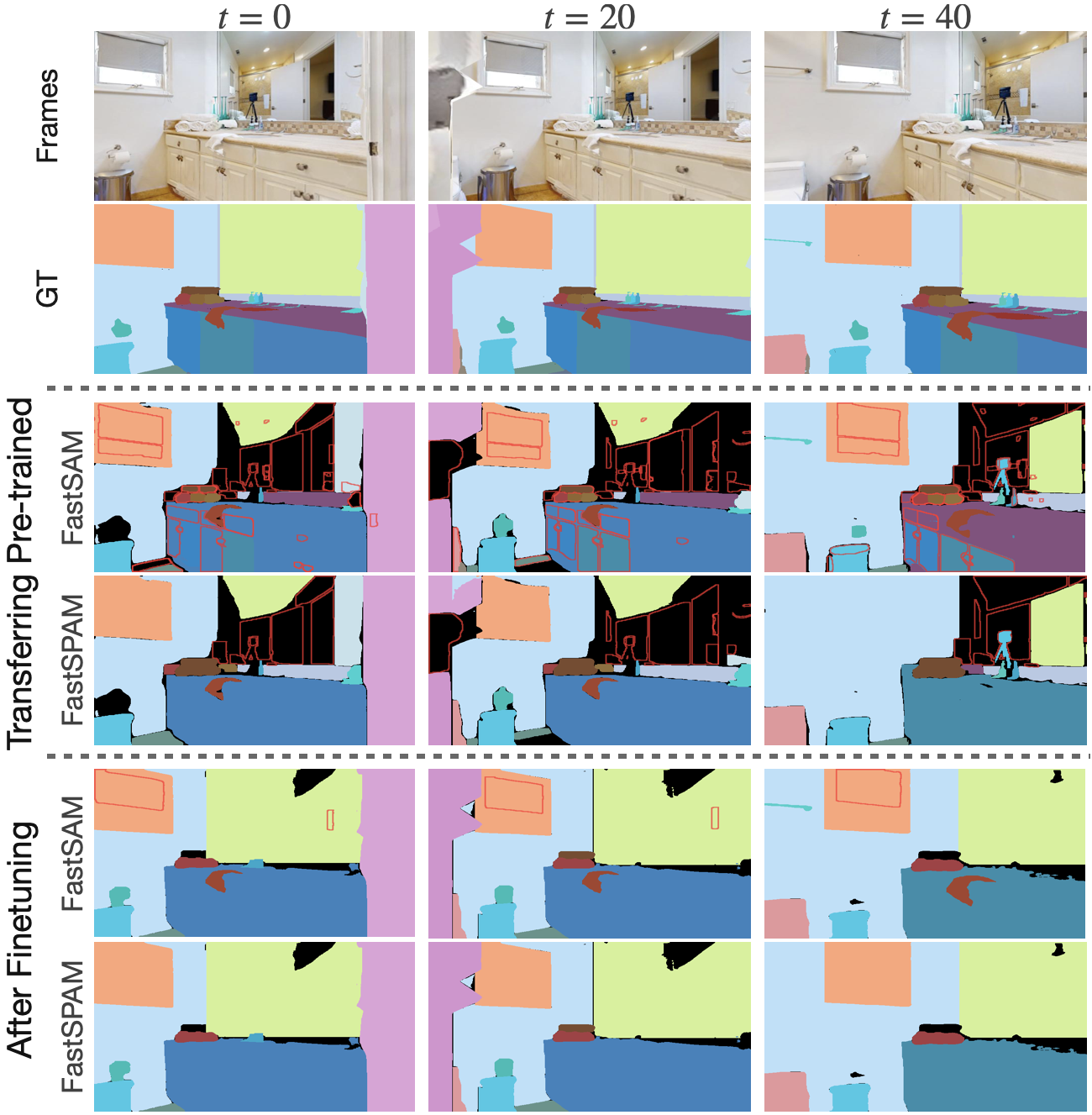}
    \vspace{-4pt}
   \caption{Qualitative comparison of FastSAM and FastSPAM. Middle panel: predictions before domain-specific finetuning. Right panel: predictions after finetuning with MVPd. FastSPAM exhibits reduced false positives as a result of self-prompting. Predicted segments are colored according to an optimal match against the ground truth segments based on pairwise F-score (\cref{sec:metric}): predicted segments matching a ground truth are assigned the corresponding color while unmatched predictions (false positives) are denoted by a red outline.}
    \label{fig:qualitative_sam_spam}
    \vspace{-8pt}
\end{figure}

Quantitative results, both in the pre-trained and finetuned settings, are shown in~\cref{tab:exp_sam_spam}.
Incorporating self-prompting resulted in an improvement of +2.55\% VSQ compared to FastSAM in the pre-trained setting and +2.73\% after finetuning.
Moreover, FastSPAM achieved higher VSQ than both pre-trained SAM models (\cref{tab:exp_pretrained}) and both finetuned baselines (\cref{tab:exp_finetuned}). 
Qualitative examples are included in~\cref{fig:qualitative_sam_spam}. 
\textbf{These results indicate that depth and camera pose features are useful features for temporally consistent segmentation and suggest future directions that make full use of these modalities during training.}

\begin{table}[t]
\vspace{-8pt}
  \centering
  \resizebox{0.925\columnwidth}{!}{
\begin{tblr}{
        colspec={Q[l]Q[c,m]Q[c]Q[c]Q[c]Q[c]Q[c,m]},
        cell{1}{1} = {r=2}{m},
        cell{1}{2} = {r=2}{m},
        cell{1}{7} = {r=2}{m},
    }
    \toprule
     Method & Training &\SetCell[c=4]{c}{{{VSQ$^k$ with $k$-Frame Window}}} & & & & VSQ\\
     \cmidrule[lr]{3-6}
    & & $k=1$ & $k=5$ & $k=10$ & $k=15$ & \\
    \midrule
    FastSAM & PT (SA-1B) & 41.02 & 31.13 & 24.19 & 20.01 & 29.09 \\
    FastSPAM & PT (SA-1B) & \textbfgray{42.95} & \textbfgray{33.65} & \textbfgray{27.05} & \textbfgray{22.89} & \textbfgray{31.64} \\
    \midrule
    FastSAM & FT (MVPd) & \textbf{61.18} & 52.03 & 44.38 & 39.09 & 49.17 \\
    FastSPAM & FT (MVPd) & 60.68 & \textbf{54.10} & \textbf{48.48} & \textbf{44.35} & \textbf{51.90} \\
    \bottomrule
\end{tblr}
}
\vspace{-4pt}
  \caption{Evaluating the impact of using depth and camera pose to form self-prompts and improve FastSAM's performance under VSQ and the MVPd test set. Models with `PT' were pretrained on SA-1B~\cite{kirillov2023segany} while `FT' models were finetuned on MVPd.}
  \label{tab:exp_sam_spam}
  \vspace{-12pt}
\end{table}

\begin{table}[b]
\vspace{-12pt}
  \centering
  \resizebox{0.95\columnwidth}{!}{
\begin{tblr}{
        colspec={Q[l]Q[c,m]Q[c,m,12mm]Q[c]Q[c]Q[c]Q[c]Q[c,m]Q[c,m]},
        cell{1}{1} = {r=2}{m},
        cell{1}{2} = {r=2}{m},
        cell{1}{3} = {r=2}{m},
        cell{1}{8} = {r=2}{m},
        cell{1}{9} = {r=2}{m},
        cell{3}{1} = {r=2}{m},
        cell{3}{2} = {r=2}{m},
        cell{3}{9} = {r=2}{m},
        cell{5}{1} = {r=2}{m},
        cell{5}{2} = {r=2}{m},
        cell{5}{9} = {r=2}{m},
        cell{7}{1} = {r=2}{m},
        cell{7}{2} = {r=2}{m},
        cell{7}{9} = {r=2}{m},
        cell{9}{1} = {r=2}{m},
        cell{9}{2} = {r=2}{m},
        cell{9}{9} = {r=2}{m},
        hline{5}={1-Z}{fg=gray9},
        hline{9}={1-Z}{fg=gray9}
    }
    \toprule
     Method & Training & Sensor Placement &\SetCell[c=4]{c}{{{VSQ$^k$ with $k$-Frame Window}}} & & & & VSQ & ${\Delta}\text{VSQ}$\\
     \cmidrule[lr]{4-7}
    & & & $k=1$ & $k=5$ & $k=10$ & $k=15$ & \\
    \midrule
    FastSAM & PT & 1m & 42.29 & 32.43 & 25.42 & 21.13 & 30.32 & 2.79 \\
    & & 0.1m & 39.37 & 29.47 & 22.65 & 18.62 & 27.53 & \\
    FastSPAM & PT & 1m & \textbfgray{44.88} & \textbfgray{35.60} & \textbfgray{28.93} & \textbfgray{24.66} & \textbfgray{33.52} & 4.21 \\
    & & 0.1m & 40.49 & 31.21 & 24.76 & 20.76 & 29.31 & \\
    \midrule
    FastSAM & FT & 1m & \textbf{62.36} & 53.31 & 45.66 & 40.35 & 50.42 & 2.78 \\
    & & 0.1m & 59.71 & 50.46 & 42.82 & 37.57 & 47.64 & \\
    FastSPAM & FT & 1m & 61.14 & \textbf{54.58} & \textbf{48.99} & \textbf{44.87} & \textbf{52.39} & 1.09\\
    & & 0.1m & 60.10 & 53.51 & 47.85 & 43.73 & 51.30 & \\
    \bottomrule
\end{tblr}
}
    \vspace{-4pt}
  \caption{Evaluating the impact of sensor placement on FastSAM and FastSPAM VSQ on the MVPd test set. Models with `PT' were pretrained on SA-1B~\cite{kirillov2023segany} while `FT' models were finetuned on MVPd. $\Delta$VSQ measures difference in model's VSQ between evaluation videos at 1m and 0.1m.}
  \label{tab:exp_sam_spam_control}
\end{table}

\vspace{-2pt}
\subsection{Controlling for Sensor Placement}
\label{sec:exp_control}
\vspace{-2pt}
This experiment set out to quantify how class-agnostic video instance segmentation models are impacted by a robot's embodiment. 
To carry out this experiment, we used the sensor placement control data in MVPd to evaluate the top-performing algorithms' VSQ performance as a function of camera height above the floor.
The pre-trained and finetuned FastSAM and FastSPAM models were evaluated on MVPd's test set videos, in which each video trajectory was recorded once at 1m height and a second time at 0.1m height.
Thus, sensor height is the controlled variable.

Quantitative results are shown in~\cref{tab:exp_sam_spam_control} indicating that in both settings and for both FastSAM and FastSPAM, lower sensor placement is consistently associated with reduced video segmentation quality.
This observed relationship may be a consequence of the distribution of camera perspectives represented in the pre-training dataset (SA-1B~\cite{kirillov2023segany}), which was captured by human photographers whose height is likely closer to 1m than 0.1m.
These results show that training on MVPd reduces the gap in performance associated with sensor height:
After finetuning on MVPd, FastSAM's performance gap as a function of sensor placement reduces from 2.79\% to 2.78\% VSQ.
The gap is further reduced by both finetuning and using self-prompting as shown by FastSPAM's gap reducing from 4.21\% to 1.09\% VSQ.
\textbf{These results suggest two directions for future work to improve video segmentation quality for embodied robots. First, temporal aggregation strategies beyond self-prompting may be useful for both models trained from supervision and finetuned models. Second, for embodied robots seeking to use video segmentation models trained from supervision, having access to data that represents the robot's morphology is beneficial and motivates using a data generation pipeline like the one used for MVPd to create embodiment-specific training data.}

\subsection{Results on Zero-Shot Subset}
\label{sec:exp_zero_shot}
\vspace{-2pt}
Next, models are compared based on their accuracy and consistency using MVPd's zero-shot subset (\cref{sec:zsmvpd}).
This comparison aims to establish a measure of accuracy to expect of models deployed in the `open-world' where they encounter categories of objects not seen at training.
Quantitative results are shown in~\cref{tab:exp3} indicating FastSPAM outperforms the top-performing baseline models as well as FastSAM on zero-shot objects by substantial margins of +6.48\%, +9.88\%, +4.38\%, and +4.69\% VSQ respectively.
\textbf{These results together with those in~\cref{sec:exp_depth_pose} indicate that self-prompting is beneficial for improving model performance on out-of-training-distribution data.}

\begin{table}[t]
  \centering
  \resizebox{0.9\columnwidth}{!}{
\begin{tblr}{
    width=\columnwidth,
        colspec={Q[l]Q[c]Q[c]Q[c]Q[c]Q[c]Q[c,m]},
        cell{1}{1} = {r=2}{m},
        cell{1}{2} = {r=2}{m},
        cell{1}{7} = {r=2}{m}
    }
    \toprule
     Method & Backbone & \SetCell[c=4]{c}{{{VSQ$^k$ with $k$-Frame Window}}} & & & & VSQ\\
     \cmidrule[lr]{3-6}
    & & $k=1$ & $k=5$ & $k=10$ & $k=15$ & \\
    \midrule
    Video K-Net & ResNet50 & 4.52 & 4.82 & 4.00 & 3.29 & 4.16 \\
    Video K-Net & Swin-base & 5.22 & 5.63 & 4.78 & 3.92 & 4.89 \\
    Tube-Link & ResNet50 & 2.28 & 1.03 & 0.82 & 0.73 & 1.22 \\
    Tube-Link & Swin-large & 2.76 & 1.26 & 1.02 & 0.94 & 1.49 \\
    OV2Seg & ResNet50 & 4.46 & 5.45 & 6.21 & 6.83 & 5.74 \\
    OV2Seg & Swin-base & 5.37 & 6.65 & 7.59 & 8.33 & 6.99 \\
    FastSAM & YOLOv8 & 14.38 & 6.50 & 3.49 & 2.36 & 6.68 \\
    FastSPAM & YOLOv8 & \textbf{22.51} & \textbf{11.96} & \textbf{6.57} & \textbf{4.43} & \textbf{11.37} \\
    \bottomrule
\end{tblr}
}
    \vspace{-4pt}
  \caption{Evaluating finetuned models on the zero-shot subset of MVPd. All objects in this evaluation belong to categories and videos that were never seen during training.}
  \label{tab:exp3}
  \vspace{-10pt}
\end{table}

\section{Conclusion}

This paper makes three central contributions: (1) the introduction of a massive RGB-D video segmentation dataset and associated pipeline to support research on embodied video segmentation, (2) extensive benchmarking experiments that establish expected performance on class-agnostic video segmentation by state-of-the-art models, and (3) ablation experiments that demonstrate depth and camera pose modalities can benefit video segmentation accuracy and consistency.
Specifically, the experiments demonstrated that incorporating spatio-temporal self-prompting (FastSPAM) with a foundation segmentation model (FastSAM) led to reduced segment inconsistency and increased model accuracy. Furthermore, the explicit self-prompting mechanism is beneficial when applied to out-of-distribution data.
Results from this study suggest future directions to further improve video segmentation models by incorporating self-prompting, depth, and camera pose during training.
Furthermore, the data contribution may enable new research directions in embodied class-agnostic video segmentation for robotic use cases.

\vspace{-2pt}

\bibliographystyle{IEEEtran}
\bibliography{main}

\end{document}